# DynMat, a network that can learn after learning

Jung Hoon Lee[1]

[1]Allen Institute for Brain Science 615 Westlake Ave N, Seattle, WA 98109 USA

Corresponding author:

Jung Hoon Lee, giscard88@gmail.com or jungl@alleninstitute.org.

## Abstract

To survive in the dynamically-evolving world, we accumulate knowledge and improve our skills based on experience. In the process, gaining new knowledge does not disrupt our vigilance to external stimuli. In other words, our learning process is 'accumulative' and 'online' without interruption. However, despite the recent success, artificial neural networks (ANNs) must be trained offline and suffer catastrophic interference between old and new learning, indicating that ANNs' conventional learning algorithms may not be suitable for building intelligent agents comparable to our brain. In this study, we propose a novel neural network architecture (DynMat) consisting of dual learning systems inspired by the complementary learning system (CLS) theory suggesting that the brain relies on short- and long-term learning systems to learn continuously. Our empirical evaluations show that 1) DynMat can learn a new class without catastrophic interference and 2) it does not strictly require offline training.

# 1. Introduction

Our knowledge and skills grow gradually by gaining experience. The exact underlying mechanisms of this 'continuous' learning remain elusive, but human learning has three distinct properties. First, our learning process is accumulative. If we "unlearn" previously obtained knowledge to accept new knowledge/skills, our brain would remain static, and our intelligence would not improve over time. Second, learning new skills/knowledge does not negate our response to external stimuli. We can concentrate on reading or studying, which does not prevent us from hearing fire alarms. If learning interferes with our vigilance, learning can put us in danger. Third, our learning is fast and immediate. We can learn a lion's appearance almost immediately at a first glance, which would increase the chance of our safety and survival.

Thus, 'continuous learning' systems comparable to the brain should be able to learn without interference between old and new trainings or offline training. Furthermore, their online learning should enable them to learn rapidly without a massive number of examples. However, ANNs (Artificial Neural Networks) including deep neural networks (DNN), which require a large number of examples, are trained offline and suffer catastrophic interference; that is, "old learning" is disrupted by "new learning"; see (Ratcliff, 1990) for details. A line of studies has proposed potential remedies of catastrophic interference (Goodfellow, Mirza, Xiao, Courville, & Bengio, 2013; Parisi, Kemker, Part, Kanan, & Wermter, 2018; Robins, 1995; Sodhani, Chandar, & Bengio, 2018), and more recent studies (Koch, Zemel, & Salakhutdinov, 2015; Vinyals, Blundell, Lillicrap, Kavukcuoglu, & Wierstra, 2016) sought learning algorithms to realize rapid learning, which is commonly referred to as 'a few or one shot-learning'. However, despite these efforts, building continuous learning systems remains difficult. Moreover, catastrophic interference and a few-shot (i.e., rapid) learning have been studied separately.

Then, how does the brain learn continuously? The complementary learning system (CLS) theory proposes that the brain utilizes short- and long-term learning systems for continuous (or continual/lifelong) learning (Norman & O'Reilly, 2003; O'Reilly, Bhattacharyya, Howard, & Ketz, 2014). The short-term learning system relies on fast memory to store new experience (or examples) using sparse and non-overlapping codes (representations); see also (Parisi et al., 2018). Due to its sparse representations, the interference between old and new items (i.e., knowledge) is minimized. The information stored in the short-term learning system can be replayed into a more effective learning system which utilizes dense and overlapping codes for information storage. Earlier studies (Gepperth et al., 2016; Hattori, 2009; Shin, Lee, Kim, & Kim, 2017) showed that networks with dual memory systems can avoid catastrophic interference. We note that even dual memory systems inspired by CLS theory have focused on addressing catastrophic interference, and their contributions to rapid learning (like a few shot learning) have not been well studied.

However, fast memory (i.e., the fast encoding of the short-term memory in the CLS theory) can also help the brain learn rapidly. The aim of this study is to seek potential short-term memory units that can both store examples rapidly and learn rapidly. By 'learning' we mean that the system should be able to predict the classes of unseen examples. Inspired by theoretical studies in neuroscience suggesting that the brain can utilize synapses to store information (Choi et al., 2018; Kleim et al., 2002; Mayford, Siegelbaum, & Kandel, 2012; Mongillo, Barak, & Tsodyks, 2008), we use synapses as memory units for short-term memory system. Specifically, we imprint inputs (examples) to synapses due to an earlier study (Diehl & Cook, 2015) showing that the brain's spike-time dependent plasticity, which is thought to underlie learning capability of the brain, can imprint inputs (examples) to synapses. With examples stored in synapses, the difference between a current example and stored ones can be natively calculated and used for predicting classes of unseen examples. To address this idea (synapse-based memory natively provides predictive power), we constructed 'dynamic matching machine (DynMat)' and estimated its performance.

As our aim is to test if synapse-based memory can enable rapid learning, we focus on measuring learning power of short-term memory, while it learns a new class after learning two classes previously. This experiment is inspired by human learning of digits, which demonstrates all three aspects of continuous learning. We learn digits one-by-one, unlike ANN/DNN. Importantly, we can learn new digits with only a few examples, but by doing so, we do not forget the previously learned digits. Thus, we test if DynMat can mimic these properties of our learning. Specifically, while DynMat learns examples from the third class after learning two classes, we estimate how well it can recognize both old and new class of objects and how rapidly its accuracy on the third class (the two old classes) improves (degrades). Since DynMat's short-term learning system does not require any iterative process, its learning process is natively 'online', and thus the two estimations would be enough to determine whether DynMat can closely mimic the brain's continuous learning. Our experiments support 1) that DynMat can learn without offline training, 2) that catastrophic interference is minimal and 3) that DynMat's learning does not strictly require a large number of training examples. With these results, we propose that synapse-based memory could be essential to continuous learning machines and should be studied further.

## 2. Related Work

Despite their proven capability and utilities in a wide range of domains, ANNs' decisions/actions can be easily biased by training examples. For instance, if networks are trained by examples of 10 cats and 1 dog, they would become more sensitive to cat examples. This can be prevented by removing biases in the training set. However, when networks learn multiple tasks sequentially, biases in the training set are unavoidable, and they would be optimized for the most recent task and forget the old tasks, which is commonly referred to as catastrophic interference/forgetting (McCloskey & Cohen, 1989). In addition, a recent study (Sodhani et al., 2018) pointed out that capacity saturation can result in catastrophic interference.

Earlier studies proposed various ways to address catastrophic interference, and they fall into three categories; see (Parisi et al., 2018; Sodhani et al., 2018) for review. First, the old examples can be replayed into the networks to reduce the bias toward the new dataset. It was proposed for feedforward networks (Robins, 1995, 2004; Silver & Mercer, 2002) and extended for reinforcement learning (Isele & Cosgun, 2018). Second, synaptic weights or other training components of the networks can be selectively updated to avoid catastrophic interference. A line of studies pursued this direction (Jung, Ju, Jung, & Kim, 2016; Kirkpatrick et al., 2017; Li & Hoiem, 2018; Liu et al., 2018). A noticeable example is the 'elastic weight consolidation' algorithm that utilizes Fisher information matrix to determine weights to be protected (Kirkpatrick et al., 2017). Third, the networks can be expanded to account for new training examples without updating synaptic weights trained for old tasks. For instance, Yoon (Yoon, Yang, Lee, & Hwang, 2017) proposed dynamically expandable networks, in which neurons are added when more capacity is necessary, and Chen et al. (Chen, Goodfellow, & Shlens, 2016) proposed 'Net2Net' algorithms which transfer knowledge to a bigger (wider or deeper) network with a higher capacity.

The studies mentioned above aimed to construct artificial learning systems that can avoid catastrophic interference. Alternatively, multiple studies ability (Gepperth et al., 2016; Hattori, 2009; Shin et al., 2017) aimed to implement CLS theory explaining the brain's continuous learning. As CLS theory emphasizes the importance of short-term memory, hypothetically in hippocampus (O'Reilly et al., 2014), they focused on developing short-term memory to replay earlier examples to the long-term learning systems. DynMat is also inspired by CLS theory and has dual memory systems, but unlike the earlier studies, DynMat utilizes synapses as short-term memory and rapid online learning.

## 3. Material and methods

DynMat was implemented by Pytorch (Paszke et al., 2017), a publicly available machine learning tool box.

### 3.1. Structure of DynMat

DynMat consists of three different areas, matching layer (ML), short-term learning module (STLM) and long-term learning module (LTLM). Each neuron (i.e., a computing node) in ML is fully connected to the input layer (Fig. 1A) via one of the training examples normalized to have a unit length (Eq. 1).

$$h_{i,k}^{ML} = \sum_j w_{ij}^{ML} \frac{x_j^k}{\|\vec{x_k}\|}, where\ w_{ij}^{ML} = \frac{x_j^l}{\|\vec{x_i}\|} \tag{1}$$

, where $h_{i,k}^{ML}$ represents the input to ML neuron $m_i$ induced by $k^{th}$ example; where $x_j^k$ is the $j^{th}$ component of $k^{th}$ example; where $w_{ij}^{ML}$ represents the connection from input node $I_i$ to ML neuron $m_i$; where $x_j^k$ is the component of $j^{th}$ component of $k^{th}$ image, and $\|\vec{x_k}\|$ represents the norm of $k^{th}$ of image. All inputs to DynMat are also normalized to have a unit length (Eq. 1), and thus a synaptic input $h_i$ to a ML neuron $m_i$ is the cosine similarities between a current example and the stored ones. Whenever an example was introduced to ML, we identified ML neurons that store examples belonging to the class of the present example. If any synaptic input $h_i$ to these ML neurons is not greater than the pre-defined threshold value $\theta$, a new ML is added to ML. That is, ML stores novel examples and calculates the similarities between the present and stored examples.

STLM is a linear layer, and each ML neuron projects one of STLM neurons (out of 3 or 10 in this study) according to Eq, 2.

$$h_{i,k}^{STLM} = \sum_j w_{ij}^{STLM} g(h_{j,k}^{ML}), where\ w_{ij}^{STLM} = \begin{cases} 0, & i \neq c \\ 1, & i = c \end{cases}, g(x) = e^{-\frac{(x-1)^2}{0.1}} \tag{2}$$

, where $h_{i,k}^{STLM}$ is the input to STLM neuron $s_i$ elicited by $k^{th}$ example; where $w_{ij}^{STLM}$ represents the connection from ML neuron $m_j$ to STLM neuron $s_i$; where $c$ represents the id of class. $g(x)$ is the activation of ML neuron used for STLM. That is, in DynMat, STLM memorizes the class of examples stored in ML. When a stored example in ML is presented, it produces the maximal input ($h_i$=1) to the ML neuron $m_i$, which was added when the example was presented. Consequently, a STLM neuron connected to this ML neuron, which is determined by the class of example (Eq. 2), will produce the strongest output. With the 'winner-take-all' rule applied,

STLM retrieves the class of the present example. However, this operation can be corrupted with stochastic activations of other ML neurons. To suppress this stochastic corruption, non-linear summation of ML outputs is used to calculate STLM neuron activation functions (Eq. 2).

LTLM is a multilayer perceptron (MLP) with a single hidden layer (Fig. 1A). The default number of hidden neurons is 200 unless stated otherwise. LTLM is also connected to ML (Eq. 3) and thus trained to classify ML outputs.

$$h_{i,k}^{LTLM} = \sum_j w_{ij}^{LTLM} h_{j,k}^{ML} \tag{3}$$

, where $h_{i,k}^{STLM}$ is the input to LTLM neuron $l_i$ elicited by $k^{th}$ example; where $h_{j,k}^{STLM}$ is the input to STLM neuron $s_j$ elicited by $k^{th}$ example; where $w_{ij}^{LTLM}$ represents the connection from ML neuron $m_j$ to LTLM neuron $l_i$.

We used backpropagation to train LTLM. The error was measured by Mean-Squared estimates, which is implemented with MSELoss in Pytorch (Paszke et al., 2017). During its learning, we used two learning phases. The first phase lasted 4000 (8000 for training with CIFAR-100 dataset) epochs with the learning rate $\lambda=1e^{-4}$, and the second phase lasted 2000 epochs (4000 for CIFAR-100 dataset) with the learning rate $\lambda=1e^{-5}$. For all training of LTLM except one exception, we used 100 examples in a single batch. When we trained LTLM with examples stored in the ML, 10 examples, instead of 100, constituted a single batch.

### 3.2. Database

To evaluate the continuous learning ability of DynMat, we used three datasets, MNIST (LeCun et al., 1998), fashion-MNIST (Xiao et al., 2017) and CIFAR-100 (Krizhevsky, 2009). MNIST includes 60,000 training and 10,000 test images of handwritten digits (0-9). Each image consists of 28-by-28 8-bit grey pixels. In fact, this dataset has been actively used to evaluate continuous learning algorithms. To make sequential tasks from the MNIST dataset, 10 digits were split into multiple disjoint sets (Chaudhry, Dokania, Ajanthan, & Torr, 2018; Lee, Kim, Jun, Ha, & Zhang, 2017; Rios & Itti, 2018; Ritter, Botev, & Barber, 2018; Zenke, Poole, & Ganguli, 2017), and pixels were randomly permuted to generate independent datasets (Goodfellow et al., 2013; Kirkpatrick et al., 2017). Because the former approach can more closely mimic our learning digits, we considered MNIST dataset as 10 disjoint sets and used either 3 or 10 disjoint sets of MNIST.

The fashion-MNIST dataset was proposed as a drop-in replacement of MNIST. It directly corresponds to MNIST in terms of the number of classes, input sizes and the sizes of test and training sets. It, however, includes examples of 10 fashion items such as t-shirts and shoes, instead of handwritten digits. CIFAR-100 is the collection of 100 classes of items ranging from animals to man-made devices. Each class has 500 training and 100 test examples, each of which is a 32-by-32 color image. All examples of the MNIST and fashion-MNIST are normalized to have unit length by dividing its norm (Eq.1) before introducing to ML; for CIFAR-100, the output images of feature detectors (see below) are normalized. When we used fashion-MNIST and CIFAR-100 datasets, we randomly selected 3 classes.

### 3.3. Feature detectors for DynMat when used as an embedded learning system

Due to its structure, DynMat can be easily embedded into other systems/networks for better performance. This approach can be useful when the tasks are too complex for fully connected layer networks like DynMat. When DynMat is embedded to other networks, it serves as a final classifier and thus the other networks can be considered as feature detectors. Since He et al. (He, Zhang, Ren, & Sun, 2015, 2016) proposed the residual deep neural network (ResNet), it has been widely adopted, and its variants such as DenseNet (Huang, Liu, Van Der Maaten, & Weinberger, 2017) and ResNext (Xie, Girshick, Dollár, Tu, & He, 2017) have been proposed. Due to its simplicity and demonstrated learning ability, we selected ResNet as our default feature detector. A pre-trained ResNet, publicly available (Idelbayev, 2018), was used in our study. It is important to note that this ResNet was trained with CIFAR-10 dataset rather than CIFAR-100. For control experiments, we also tested DenseNet, ResNext and VGG networks (Simonyan & Zisserman, 2015) as feature detectors. Like ResNet, pre-trained models with CIFAR-10 in a public domain (Wang, 2019) were used.

### 4. Results

The schematics of DynMat are illustrated in Fig. 1A; it consists of matching layer (ML), short-term learning module (STLM) and long-term learning module (LTLM), which is consistent with CLS theory. ML, the first stage of DynMat, receives external inputs and forwards outputs to STLM and LTLM. The external inputs $\vec{x^k}$ to ML are normalized to have unit lengths (Eq. 1), and ML stores these normalized examples by imprinting them to synaptic connections. As a result, a synaptic input $h_i$ to a ML neuron $m_i$ is proportional to cosine similarities between a present example and examples previously stored in synapses targeting the neuron $m_i$ (Eq. 1). This means the difference between old and new examples are natively calculated by ML. The number of ML neurons is not fixed. Instead, it grows when a present input is substantially different from the previously introduced examples; that is, each ML neuron stores one example. Specifically, a new ML neuron is added if no input $h_i$ is higher than the threshold value $\theta$; that is, when a present example is substantially different from the previously stored examples. This comparison is conducted among examples within the same class. For instance, an instantiation of digit 2 is compared to other digit 2 instantiations stored in ML.

STLM is a linear layer working as a short-term learning system, in which each neuron represents (codes) a class (Fig. 1A); that is, the number of neurons in STLM is the same as the number of classes introduced to it. STLM does not need to be trained offline to classify inputs from ML neurons. Instead, each ML neuron is exclusively connected to a STLM neuron according to its class (Eq. 2), which means STLM's learning is instantaneous and intrinsically online. For instance, when an instantiation of digit '2' is imprinted to synaptic weights converging to a newly inserted ML neuron, the new ML neuron is connected to the STLM neuron representing the class (i.e., digit) '2'. Inputs to STML neurons are defined by nonlinear activations of ML neurons (Eq. 2).

LTLM corresponds to CLS theory's long-term learning system. Since it is proposed to utilize the distributed and overlapping codes, we construct LTLM using a hidden-layer perceptron known to use overlapping representation (Hertz, Krogh, & Palmer, 1991); the default number of hidden neurons is 200. In principle, there is no restriction on the structure of LTLM. Any network that can be trained offline to perform tasks effectively (e.g., classification in this study) can be used as a LTLM. We select a hidden-layer perceptron due to its well-documented learning power (Hertz et al., 1991; LeCun, Bottou, Bengio, & Haffner, 1998). As shown in Eq. 3, inputs to LTLM are the linear summations of inputs of ML neurons; that is, the gain of ML neurons for LTLM is set to 1. With ML outputs, LTLM is trained using a common backpropagation with mean-squared estimated error (see Section 3.1).

## 4.1. Empirical evaluation protocol of the continuous learning ability of DynMat

To test the continuous learning capability, we train DynMat to learn a new class after previous learning of two classes. It should be noted that STLM and ML learn new information online while we introduce the third-class objects. LTLM learns all three classes together offline after introducing the third class. While examples from the third class are introduced, we measure STLM's error rates on the classes of test examples from the two old classes and the new class. First, the error rate on the two old classes will allow us to determine if DynMat suffers from catastrophic interference. If it suffers from catastrophic interference, the error rate on the two old classes (i.e., the test examples from them) would increase rapidly while STLM learns a new class. Second, the error rate on the new third class will allow us to determine if DynMat can learn a new class rapidly. If STLM's accuracy on the third class is not improved until a large number of examples of the class is introduced, STLM cannot learn it rapidly, which means that STLM fail to reproduce the crucial property of our continuous learning.

In this study, we use three datasets (MNIST, fashion-MNIST and CIFAR-100) (Krizhevsky, 2009; LeCun et al., 1998; Xiao, Rasul, & Vollgraf, 2017); see Section 3.2.

Regardless of the datasets, we use the same protocol as follows:

1. We use all training examples of the first two classes to build ML and STLM (Eqs. 1 and 2). Then, LTLM is trained with linear outputs of ML (Eq. 3). This step is designed to generate the initial state of DynMat to test its continuous learning ability.

2. We expose DynMat to examples of the third-class and update ML and STLM accordingly (Eqs. 1 and 2). During the exposure, whenever a new ML neuron is added to ML, the error rates of STLM on test examples of all three classes are estimated.

3. We train LTLM with training examples of all three classes and test its performance on the test sets of three classes.

In most of our experiments, we use three classes of these datasets to focus on analyzing STLM's learning of new classes (Sections 4.2-4.4), but we further test the continuous learning ability of DynMat by using 10 classes (i.e., digits) from MNIST (Section 4.5); indeed, this approach has been used in earlier studies (Chaudhry et al., 2018; Lee et al., 2017; Rios & Itti, 2018; Ritter et al., 2018; Zenke et al., 2017).

Since ML is fully connected to the input layer, DynMat can replace any fully-connected networks including multi-layer perceptrons (MLPs) which have been trained to perform numerous tasks (Hertz et al., 1991). For instance, MLPs can learn to recognize handwritten digits by looking at 28-by-28 gray pixel images (LeCun et al., 1998). Recently, however, the fully-connected layer network is used as a final classifier in DNNs. That is, a fully-connected network can work as a standalone (e.g., MLPs) learning system or an embedded (e.g., the final classifier in DNNs) one. Thus, we test DynMat as both a standalone system (Fig. 1A) and an embedded system (Fig. 1B), and the results are discussed in Sections 4.2 and 4.3, respectively.

## 4.2. DynMat as a standalone learning system

We ask if DynMat can work as a standalone system similar to MLPs by training it with raw pixel images included in MNIST and fashion-MNIST.

For MNIST dataset, we use training and test examples of three digits (i.e., classes) '0',' 1' and '2'. As stated above, we first store the training examples of '0' and '1' in STLM/ML and then train LTLM with them. During this initial stage, we test the effects of the threshold value $\theta$, which is used to detect the novel inputs (see Eq. 1 and text below it), on the number of ML size and LTLM's error rates. As shown in Fig. 1C, when the bigger $\theta$ is used, more examples are stored in ML. As each ML neuron stores one example, this means that $\theta$ decides the ML size. Also, we note that LTLM can reliably recognize the examples in the test set (Fig 1C) and that its accuracy improves, as $\theta$ increases.

Next, we introduce digit '2' training examples (the third-class examples) to DynMat. When presenting each example of digit 2, the inputs to ML neurons that store examples of '2' are compared with the threshold value $\theta$ to determine whether the current example is substantially distinct from the earlier examples and needs to be stored. When a new neuron is added to ML to store a new example, it is connected to the neuron representing digit '2' in STLM; all new examples are drawn from digit 2 training examples. Whenever a new ML neuron is added, we evaluate STLM's classification error on all test examples of the two old digits (0 and 1) and the new digit (2), separately. Figure 2A shows the changes in measured error rates with $\theta=0.5$ during the introduction of the third-class. STLM01 represents the error rate on 0 and 1, whereas STLM2 represents the error rate on 2. These error rates are shown in a logarithmic scale. As more examples are presented, the number of digit 2 examples stored in ML becomes larger (see *x*-axis of Fig 2A). Figures 2B-D show the same results, but a distinct $\theta$ is used in each case. As $\theta$ becomes higher, the performance of STLM is improved. That is, STLM's accuracy improves, as ML size grows.

We make two germane observations independent of $\theta$. First, as the number of digit '2' examples stored in ML, shown in *x*-axis, increases, the error rate of STLM on the new digit '2' declines. Second, the error rate of STLM on two digits (0 and 1), which was previously trained, rises. The increasing error rate on the two digits (0 and 1) can be the result of the interference between old and new learning, but it should be noted that this increasing error rate is quite limited (Figs. 2A-D). Importantly, the error rate on test examples of digit 2 decreases rapidly, especially when a small number of training examples of digit 2 are stored. These results suggest 1) that STLM learns digit 2 without forgetting the two old digits 0 and 1 and 2) that it can learn new class objects (i.e., 2) using a small number of examples; once again, the error rates of STLM are calculated using test examples, not training examples.

After seeing all available examples of digit 2, STLM recognizes all three examples at a low error rate (less than 1 % with $\theta$=0.8, shown in Fig. 2D). As expected, the number of examples of all three digits stored in ML increases, as $\theta$ increases (Fig. 2E). Finally, we train LTLM with all three digits and compare the error rates on the test set between LTLM and STLM (Fig. 2F). STLM produces more errors, but the difference between STLM and LTLM becomes smaller, as $\theta$ increases. It is rather unexpected that STLM and LTLM error rates are nearly equivalent, even though STLM is not trained offline. This may be attributed to the simplicity of MNIST dataset.

Thus, to further test DynMat as a standalone learning system, we also train DynMat with the fashion-MNIST, a proposed drop-in replacement of MNIST (Xiao et al., 2017). Fashion-MNIST includes images of 10 fashion items such as shoes (see Section 3.2). After randomly selecting 3 items out of 10 items in the dataset, we repeat the same experiments above. Figures 3A-D show the error rates of STLM on test examples depending on various threshold values during the exposure of the third-class. STLM produces more errors on fashion items (fashion-MNIST) than on handwritten-digits (MNIST), but we observe the same trend. First, the classification error on the third-class object improves rapidly with minimal increase in the error rate on the test examples of the first two classes (Figs. 3A-D); again, the *x*-axis represents the number of examples of the third-fashion item stored in ML. Second, the difference between STLM and LTLM becomes smaller, as $\theta$ increases. That is, STLM can learn continuously. Finally, we further test DynMat by conducting 10 independent experiments, in which three fashion items are randomly chosen. In all experiments, the ML size increases (Fig. 3E), and STLM's classification fidelity improves (Fig. 3F), as $\theta$ increases. As $\theta$ controls the size of ML layer (i.e., the size of input layer of STLM), this means that STLM's accuracy improves as it stores more examples.

These results suggest 1) that the STLM in DynMat can learn a new class after learning other classes without offline training or catastrophic interference and 2) that even without offline training, STLM can learn new class objects with a small number of examples. In addition, LTLM can be re-trained with all classes including the new one and provide a better classification fidelity when DynMat can afford offline training.

### 4.3. DynMat as an embedded learning system

In recent breakthroughs in deep learning (Lecun, Bengio, & Hinton, 2015; Schmidhuber, 2015; Vargas, Mosavi, & Ruiz, 2017), fully-connected networks are embedded into deep neural networks to classify outputs of convolutional networks (ConvNets); the outputs of ConvNets are often referred to as features of visual images (or input vectors). Given the explosive applications of deep neural networks, it is important to test if DynMat can work with the features detected by ConvNets. To this end, we use residual deep neural networks (ResNet) proposed by (He et al., 2015) as feature detectors. Specifically, we replace the fully connected layer in ResNet with DynMat (Fig. 1B) and test the classification error of DynMat using the same experimental protocol. In this experiment, we use the 'ResNet44' publicly available (Idelbayev, 2018); ResNet44 will be referred to as ResNet hereafter. It should be noted 1) that the ResNet is trained with CIFAR-10 dataset instead of CIFAR-100 used to test DynMat and 2) that we test DynMat using the three visual objects (i.e., classes) randomly chosen from CIFAR-100. If feature detectors (i.e., ResNet in this study) are trained with CIFAR-100 dataset, inputs to DynMat (which replace the fully-connected layer in the original ResNet) may be highly optimized for classification, and the performance of DynMat as a continuous learning system can be overestimated. To avoid this potential bias, we use pretrained ResNet with CIFAR-10 dataset, instead of CIFAR-100.

As with experiments with MNIST and fashion-MNIST, we measure the classification error of STLM on test examples during the exposure of the third-class objects (Figs. 4A-D). The increase in error rate on the two previously trained class objects is much slower than the reduction of error rate on the third-class object, suggesting that STLM learns the features detected by ResNet without offline training or catastrophic interference, as discussed above. We also perform 10 independent experiments, in which 3 visual classes are randomly drawn out of 100 classes. In all experiments, as $\theta$ increases, the number of stored examples increases (Fig. 4E), and the classification error of STLM approaches that of LTLM (Fig. 4F), supporting that DynMat can work as an embedded learning system as well as a standalone system.

To further test the learning power of DynMat as an embedded learning system, we test alternative feature detectors by replacing ResNet with DenseNet, ResNext and VGG networks trained with the CIFAR-10 dataset. With these alternative feature detectors, we repeat the protocols above to train DynMat and estimate its error rates on randomly selected 3 image classes from CIFAR-100 database. Figure 5 shows the error rates of the DynMat with all three feature detectors during the exposure to the third class. We note that DynMat's accuracy depends on the feature detectors: DynMat shows the lowest accuracy when VGG19 is used as a feature detector. Despite this precision variability, however, DynMat can continuously learn with all these three feature detectors. DynMat's accuracy on the third class (shown in blue line in Fig. 5A, B and C) improves fast especially when a small number of examples are stored in ML, whereas the degradation of its accuracy on the two first two classes (shown in red line in Fig. 5A, B and C) is limited. We also find that regardless of feature detectors the error rates of STLM and the gap between the accuracy between LTLM and STML decrease (Fig, 5D, E and F), as the threshold $\theta$ increases; once again, $\theta$ controls the size of ML.

### 4.4. DynMat as a self-contained learning system

So far, we have discussed the classification error of LTLM trained with all available training examples to estimate the upper bound of LTLM's classification fidelity. However, as the offline training of LTLM requires an external memory storage of training examples, it could be expensive for agents to accumulate experience over a long period. As DynMat stores a subset of examples in synapses in ML, the stored examples can be used to train LTLM to remove external memory. Thus, we ask if the examples stored in ML can suffice to train LTLM to perform reliable classification. Since the number of examples in ML depends on $\theta$, we measure the classification error of LTLM trained with examples stored in ML depending on $\theta$. Specifically, we compare a LTLM trained with CIFAR-100 examples stored in ML against a LTLM trained with the full training set. Additionally, we compare the classification error of LTLM to that of STLM. We note 1) that the accuracy of LTLM (shown in blue and cyan in Fig. 6A), which is trained with stored examples in the ML, improves and becomes closer to that of LTLM trained with the full training set (shown in red and green in Fig. 6A), as $\theta$ becomes higher and 2) that the classification error of LTLM trained with stored examples in the ML becomes better than that of STLM (Fig. 6B), when $\theta \geq 0.65$. These results indicate that LTLM's accuracy can be largely maintained when it is trained with internally stored examples in DynMat. We further test the performance of LTLM trained with examples of fashion-MNIST and find equivalent results (Fig. 6C and D). Based on these results, we propose that the offline training of LTLM does not strictly require external memories.

## 4.5. Scalability of DynMat

The results presented above describe the continuous learning ability of DynMat for three classes. However, continuous learning systems are expected to learn more than three classes and tasks. Thus, to test if DynMat's continuous learning ability is scalable and can be extended to learn more than 3 classes, we train DynMat to learn all 10 digits included in the MNIST dataset sequentially. That is, we use disjoints of MNIST, as used in earlier studies (Chaudhry et al., 2018; Lee et al., 2017; Rios & Itti, 2018; Ritter et al., 2018; Zenke et al., 2017). As before, we first train two digits 0 and 1 and introduce 8 digits (2-9) sequentially to DynMat and enable its STLM/ML to learn them. Figure 7A and B shows the accuracy of STLM when DynMat is exposed to digits 4 and 9, respectively. In these figures, the blue lines represent the error rates of STML on digit 4 (Fig. 7A) and on digit 9 (Fig. 7B), whereas the red lines represent the error rates of STML on previously trained digits, 0-3 in Fig 7A and 0-8 in Fig. 7B. In both cases, the error rates on new digits decrease rapidly, and the error rates on the old digits increase marginally; these error rates are shown in a logarithmic scale. We find the same trend during STLM's learning of all 8 digits. In Fig 7C, we summarize the error rates of STLM on previously learned digits during DynMat's exposure to one of 8 digits (2-9). The color codes represent the new digits that are newly introduced to DynMat. That is, the red line in Fig 7C represents STML's error rates on digits 0-3, while DynMat learns digit 4. Also, the error rates of STML on new digits decrease rapidly, as shown in Fig 7D. In Fig 7D, the same color codes are used to specify the newly introduced digits; for instance, the red line represents the error rate of STLM on digit 4 (not digits 0-3). Even with all 10 digits trained, STLM shows slightly worse accuracy than LTLM (with 500 hidden neurons), as shown in the inset of Fig 7E; their error rates are far below the error rate reported by the original study (LeCun et al., 1998). These results suggest that DynMat's continuous learning ability is not limited to the three classes and can be extended further.

Next, we examine the effects of presentation orders of classes on STLM's accuracy by shuffling sequences of digits (i.e., classes) introduced to DynMat. We also shuffle presentation orders of the same digits; for instance, examples of '4' are presented in random orders across experiments. We conduct 10 experiments, in which the sequence of digits is independently shuffled. Figure 8 shows three experiments. In the experiment shown in Fig 8A and B, digit 4 is presented after presenting 2, 9, 6, 4, 0, 3, 1, 7, 8 and 5. Fig. 8A shows the STML's error rates on digit 4 and previously trained digits (2,9,6), and Fig 8B shows the error rates of STLM on new digits during their exposure. As shown in these figures, STLM exhibits the qualitatively same behavior as before regardless of the presentation order. In other two experiments with randomly shuffled sequences, we find consistent results (Fig. 8C-F).

If presentation orders of examples are shuffled, STLM stores different examples due to its operating principle. Even with different examples stored in STLM, the final performance of STLM on all 10 digits are almost equivalent over 10 experiments (Fig. 9A). As shown in Fig. 9A, the variability of STML error rates across 10 experiments is in the order of 0.1%; the error rates of LTLM show much higher variability. Finally, we test the dependency of LTLM's learning power on the hidden layer size. In doing so, we use the identical sequence of digits and measure LTLM's error rates with 10 different hidden layer sizes (from 100 to 1000). As the same sequence is used, STLM shows identical accuracy, but LTLM's accuracy increases, as the hidden layer size grows; with 100 hidden neurons, LTLM fails to learn all 10 digits (Fig. 9B).

## 5. Discussion

In our empirical experiments, we made two germane observations. First, while STLM in DynMat learns a new class, it does not forget the previously trained class objects. That is, STLM does not suffer from catastrophic interference. Second, STLM's prediction on unseen (i.e., test) examples of a new class improves rapidly, as the number of new class training examples stored in ML increases. This observation indicates that STLM does not require a massive number of training examples and that STLM may be able to perform a few-shot learning, especially when a desired task does not strictly require high accuracy. Together with STLM's instant online learning capability, the two observations above suggest that synapse-based memory can help learning machines learn rapidly from a small number of examples, even without offline training, and avoid catastrophic interference.

## 5.1. The operations of DynMat and their links to earlier studies

Synapse-based memory in DynMat makes its learning 'online' and 'fast' and allows it to avoid catastrophic interference. First, with inputs (i.e., examples) imprinted to synapses, DynMat can natively perform 'matching' operation and thus infer which old example is the closest to the current input. This matching operation makes STLM predict the class (i.e., the label) of unseen examples precisely even without formal learning processes. We note that the matching between examples was used for one-shot learning or a few-shot learning (Koch et al., 2015; Vinyals et al., 2016). Second, ML expands its size by adding more synapses, when its capacity is saturated; in fact, the number of (sets of) synapses represents the capacity of ML. That is, in principle, DynMat belongs to the 'dynamic architectures' approach for continuous learning machine (Parisi et al., 2018).

Due to the importance of short-term memory in CLS theory, the earlier studies in its neural implementations sought effective substrates of short-term memory. To the best of our knowledge, no other theory proposes synapses as potential substrates of short-term memory. Further, our results suggest that synapse-based memory could make short-term memory not just store temporary information but also make prediction on unseen inputs.

## 5.2. Potential extension of DynMat for recurrent networks

Continuous learning is essential in building human-level intelligence, but it also has practical importance in deploying ANNs to real world problems. Our environment constantly changes over time, and training examples cannot fully capture inputs to ANNs in the real world. For instance, a security camera can be trained to detect pedestrians' specific behavior in multifarious weather conditions, but it is impossible to perform training with all weather conditions, which may cause failures and/or necessitate further training. With continuous learning ability, a security camera could automatically compensate for differences between training and actual conditions. Furthermore, we note that most real-world problems have temporal dynamics, for which feedforward networks cannot account. As recurrent networks are developed to predict the temporal dynamics and correlations (Hertz et al., 1991), recurrent networks with continuous learning ability would be essential for deploying ANNs to real-world problems.

Then, can DynMat be applied to recurrent networks? It could be possible to use DynMat as a reading layer for Echo or Liquid state machines, which have been successfully applied to multiple real-world problems (Jaeger, 2007; Tanaka et al., 2018). Liquid (or Echo) state machines (LSM) generate dynamically changing patterns which are read out by linear layer to perform computations (Jaeger, 2007; Maass, Natschl, & Markram, 2003). As DynMat can replace any fully connected linear networks, it will be straightforward to couple it with LSM or its variants. With this extension, DynMat may allow recurrent networks that are trained for real-world problems to automatically adapt to actual environments that are different from training environments.

## 5.3. Potential extension of DynMat towards brain-like machine

Based on our encouraging results, we plan to extend DynMat to mimic the brain's functions more closely. Specifically, we will seek potential algorithms for DynMat to mimic prefrontal cortex (PFC) which is known to perform executive functions such as decision-making (Miller & Cohen, 2001). To this end, DynMat will be extended in two ways.

First, we will develop unsupervised and reinforcement learning algorithms for DynMat (Hertz et al., 1991), as PFC is known to be associated with reward-based learning (Duverne & Koechlin, 2017; Miller & Cohen, 2001). The current DynMat requires labeled examples, but is it possible for DynMat to learn a new task without them? In principle, DynMat can be modified to learn without labeled examples, if two changes are introduced. First, with labeled examples, ML neurons are inserted when the present example is substantially different from the stored examples of the same class. However, if labels of examples are not available, comparison within a class would not be possible. That is, the comparison should be agnostic to the classes. Second, LTLM needs to be a network that is trained with rewards, as suggested by reinforcement learning (Hertz et al., 1991). Since there is no fundamental restriction on LTLM structure, adopting one of the reinforcement learning algorithms to LTLM is possible.

Second, we will develop algorithms to regulate reciprocal interactions between feature detectors and DynMat, as PFC is reciprocally connected with sensory and motor areas (Bedwell, Billett, Crofts, & Tinsley, 2014). In this study, the feature detectors (ConvNets in ResNet), which remain static during DynMat operation, provide afferent inputs to DynMat, but it does not receive afferent inputs from DynMat. This is consistent with the notion that high-order cognitive areas such as PFC perform executive functions based on features extracted by low-order sensory cortices. However, it is increasingly clear that reciprocal interactions between low-order sensory and high-order cognitive areas are critical in the brain's cognitive functions; see (Bastos et al., 2015; Buschman & Miller, 2007; Fries, Reynolds, Rorie, & Desimone, 2001) for instance. Therefore, we will investigate the algorithms to establish reciprocal interactions between feature detectors and DynMat to improve DynMat's learning ability.

# 6. Acknowledgement

JHL wishes to thank the Allen Institute founders, Paul G. Allen and Jody Allen, for their vision, encouragement and support.

*Acad Sci U S A*, *99*(20), 13228–13231. http://doi.org/10.1073/pnas.202483399

Koch, G., Zemel, R., & Salakhutdinov, R. (2015). Siamese Neural Networks for One-Shot Image Recognition. In *ICML* (Vol. 37).

Krizhevsky, A. (2009). Learning Multiple Layers of Features from Tiny Images. *Technical Report, University of Toronto*, 1–60. http://doi.org/10.1.1.222.9220

Lecun, Y., Bengio, Y., & Hinton, G. (2015). Deep learning. *Nature*, *521*(7553), 436–444. http://doi.org/10.1038/nature14539

LeCun, Y., Bottou, L., Bengio, Y., & Haffner, P. (1998). Gradient-Based Learning Applied to Document Recognition. *PROC. OF IEEE*.

Lee, S.-W., Kim, J.-H., Jun, J., Ha, J.-W., & Zhang, B.-T. (2017). Overcoming Catastrophic Forgetting by Incremental Moment Matching. In *NIPS*. http://doi.org/10.3857/roj.2012.30.1.36

Li, Z., & Hoiem, D. (2018). Learning without Forgetting. *IEEE Transactions on Pattern Analysis and Machine Intelligence*, *40*(12), 2935–2947. http://doi.org/10.1109/TPAMI.2017.2773081

Liu, X., Masana, M., Herranz, L., Van de Weijer, J., Lopez, A. M., & Bagdanov, A. D. (2018). Rotate your Networks: Better Weight Consolidation and Less Catastrophic Forgetting. *ArXiv*, 1802.02950. http://doi.org/arXiv:1802.02950v3

Maass, W., Natschl, T., & Markram, H. (2003). Computational Models for Generic Cortical Microcircuits A Conceptual Framework for Real-Time Neural Computation, 1–26.
28

# Figures

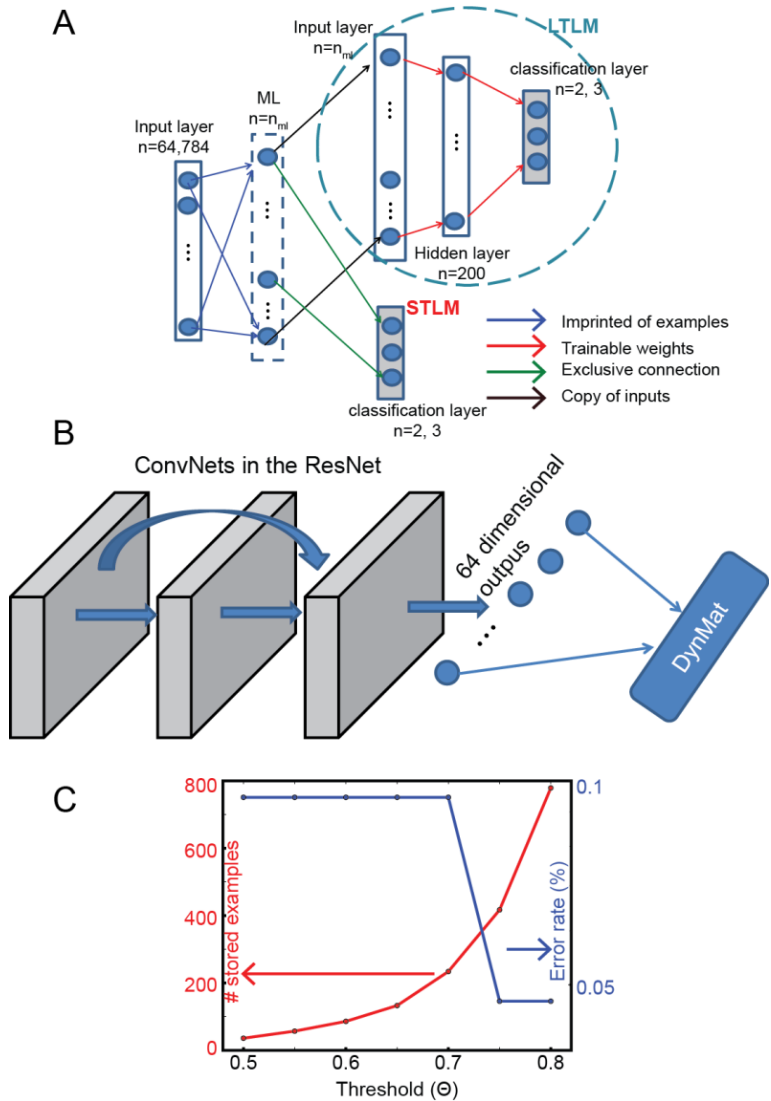

**Figure 1: Structure of DynMat and basic mechanism. (A)**, The schematics of DynMat consisting of matching layer (ML), short-term learning module (STLM) and long-term learning module (LTLM). In principle, DynMat is a collection of layered neurons (i.e., the computing nodes). The size of each layer is shown in the figure, and the input size is 784 for MNIST and fashion-MNIST and 64 for CIFAR-100 datasets. The arrows represent synaptic weights, and their properties are summarized in color codes. See Section 3 for details. **(B)**, The schematics of DynMat embedded into ResNet. The ResNet adopted from the public repository (Idelbayev, 2018) generates 64 dimensional outputs, which are fed to the linear classifier. In this study, the linear classifier is replaced with DynMat **(C)**, The dependency of ML size and the error rate of LTLM on the threshold ($\theta$). Specifically, the size and error rate are measured using examples of two digits (0 and 1).

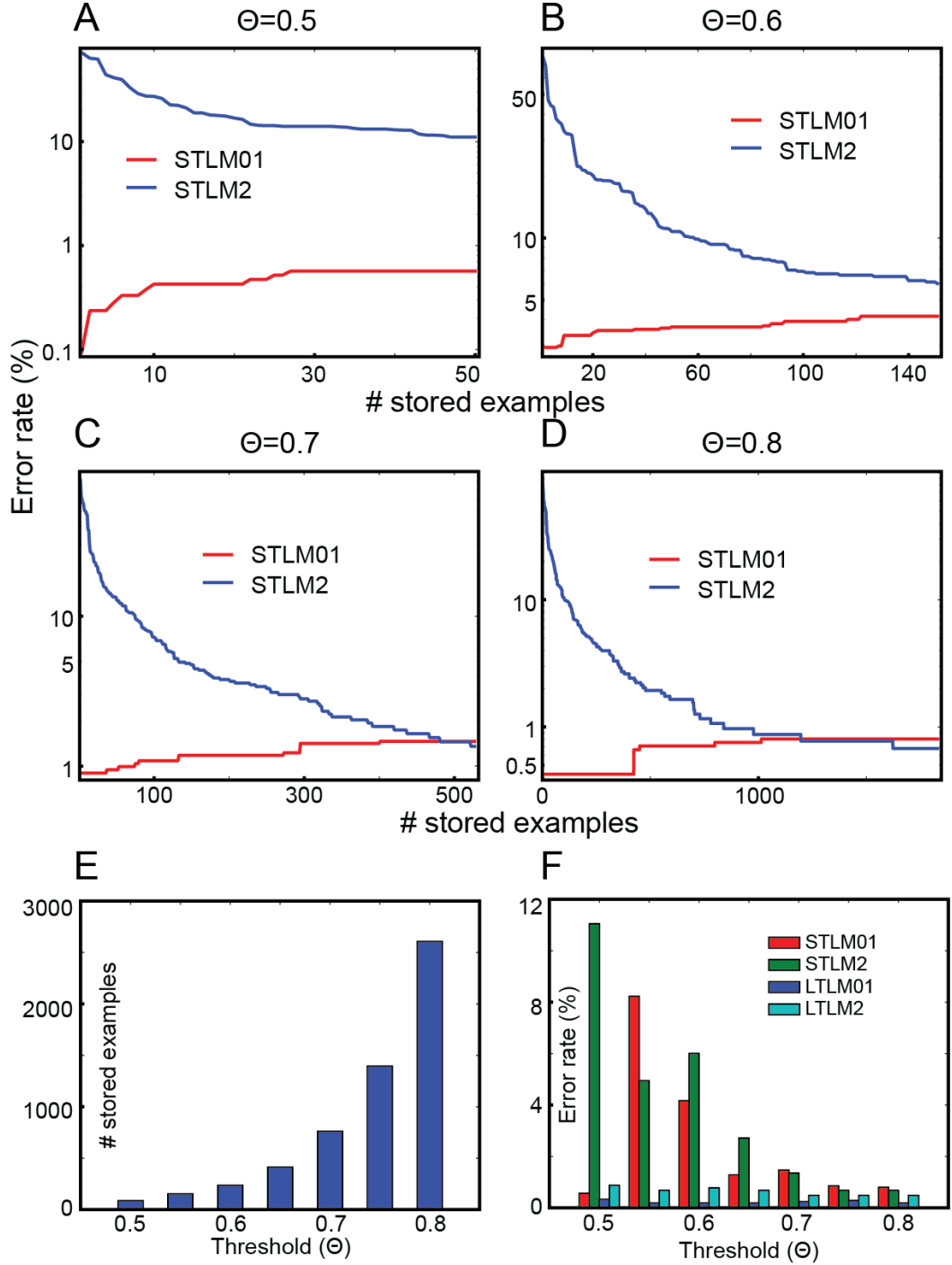

**Figure 2: DynMat's learning with MNIST dataset**. **(A)-(D)**, The error rate of STLM, during the exposure of the examples of new digit 2, on the first two testing examples (digits 0 and 1) and the third example (digit 2). STLM01 represents the error rate on 0 and 1, whereas STLM2 represents the error rate on 2. These error rates are shown in a logarithmic scale. These error rates depend on the threshold ($\theta$), and thus we measured them by varying the threshold from 0.5 to 0.8. The results with $\theta$=0.5, 0,6, 0.7 and 0.8 are shown in (A), (B), (C) and (D), respectively. The *x*-axis represents the number of examples of digit 2 stored in ML (i.e., the number of ML neurons). **(E)**, The total number of examples stored in ML depending on the threshold value ($\theta$). **(F)**, The error rate of STLM and LTLM. LTLM01 (STLM01) represents the error rate of LTLM (STLM) on digits 0 and 1, and LTLM2 (STLM2) represents the error rate of LTLM (STLM) on digit 2; all error rates are estimated using the test set.

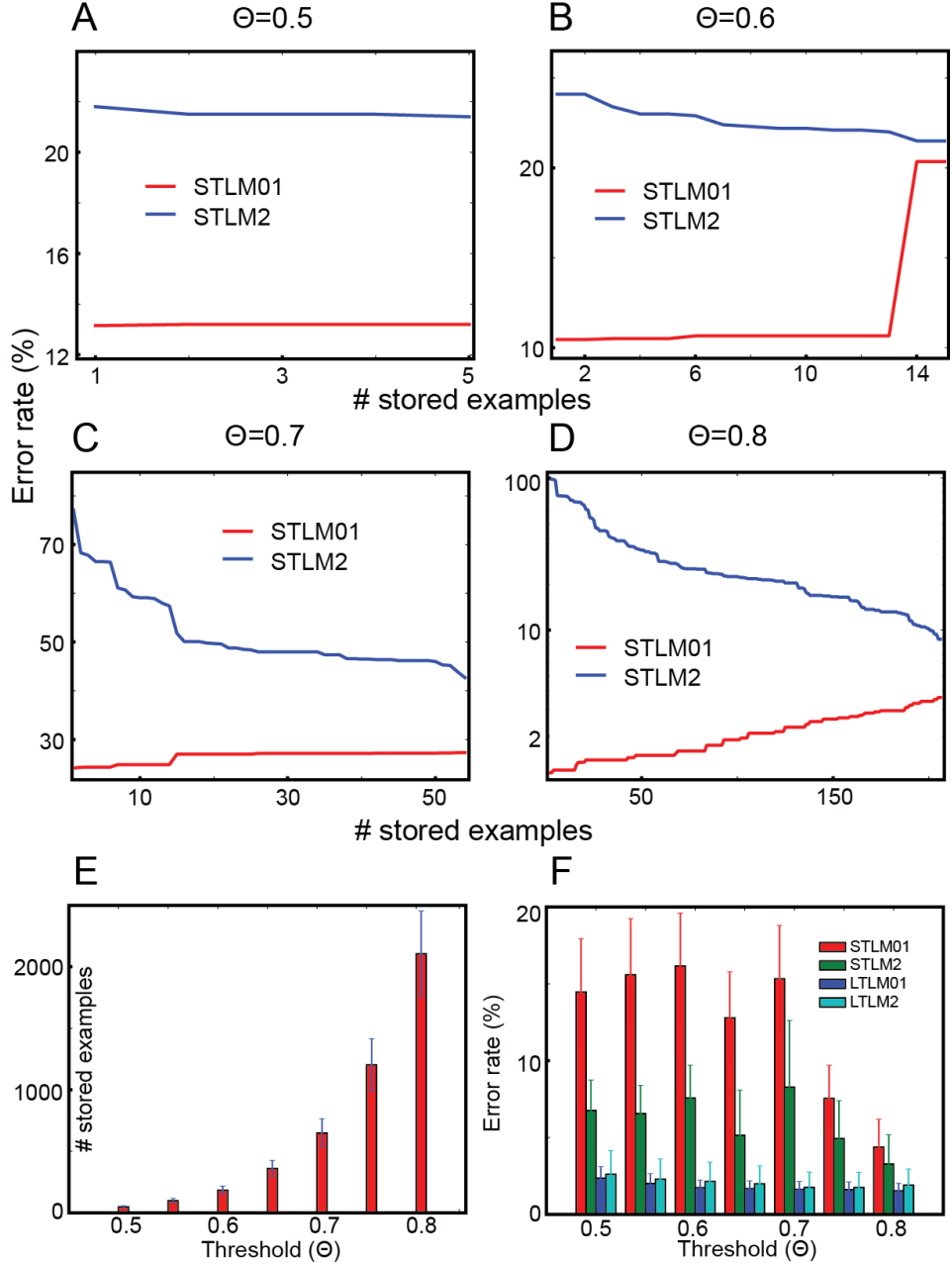

**Figure 3: DynMat's learning with fashion-MNIST dataset**. **(A)-(D)**, The error rate of STLM, during the exposure of examples of the third item, on the testing examples of the first two fashion items and the third fashion item; they are selected randomly out of 10 items in the dataset. STLM01 represents the error rate on the first two items, whereas STLM2 represents the error rate on the third item. These error rates depend on the threshold ($\theta$), and thus we measured them by varying the threshold from 0.5 to 0.8. The results with $\theta$=0.5, 0,6, 0.7 and 0.8 are shown in (A), (B), (C) and (D), respectively. The *x*-axis represents the number of examples of third-fashion item stored in ML. **(E)**, The total number of examples stored in ML depending on the threshold value ($\theta$). We illustrated the mean values and standard errors calculated from 10 independent experiments, in which 3 fashion items are independently chosen; the results suggest that the variance is high. **(F)**, The error rate of STLM and LTLM. LTLM01 (STLM01) represents the error rate of LTLM (STLM) on the first two items, and LTLM2 (STLM2) represents the error rate of LTLM (STLM) on the third item. We illustrated the mean values and standard errors calculated from 10 independent experiments. All errors are estimated using the test set, and logarithmic scales are used in *y*-axis in (D).

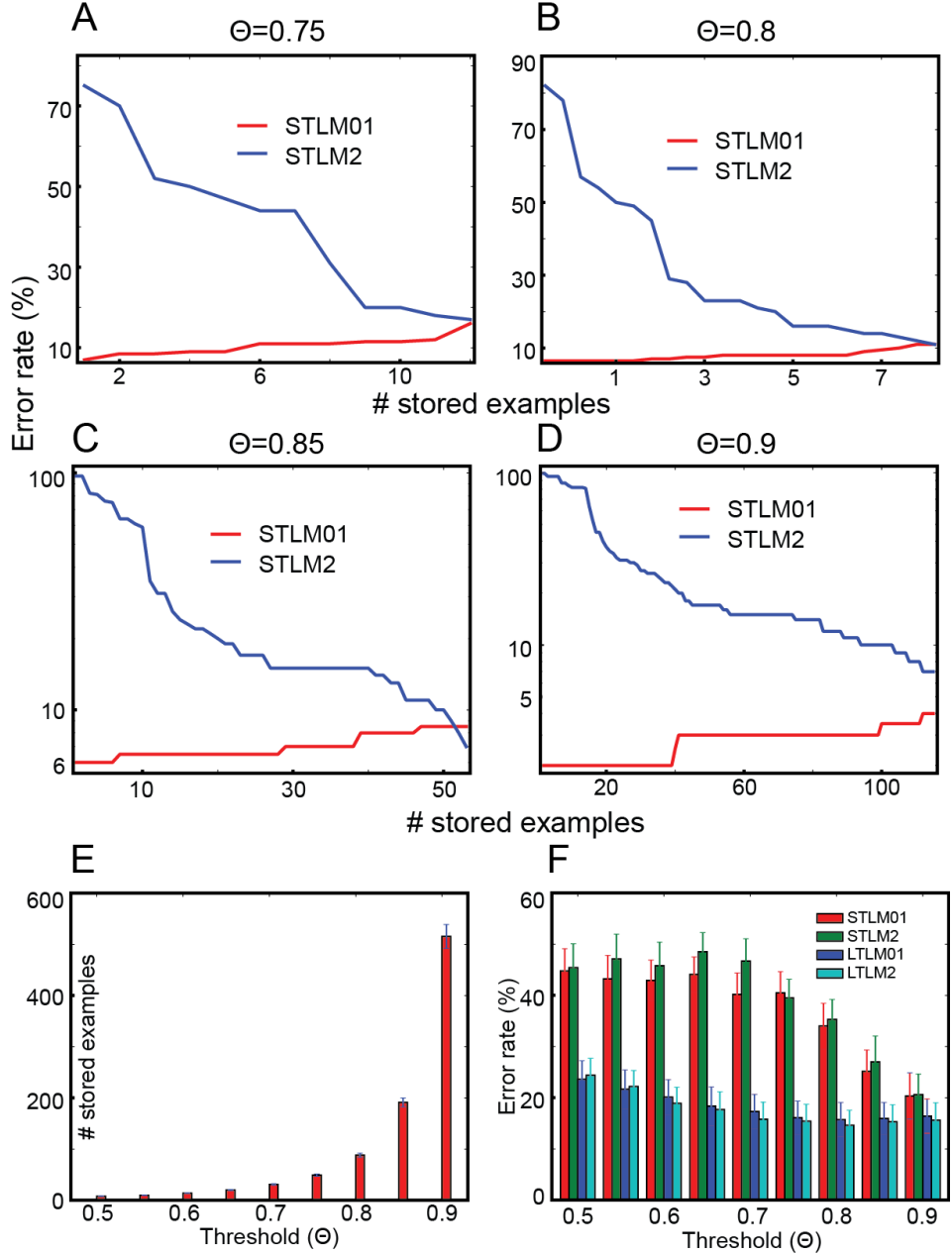

**Figure 4: DynMat's learning as an embedded learning system**. **(A)-(D)**, The error rate of STLM, during the exposure of examples of the third item, on the testing examples of the first two visual items and the third-visual object item; they are selected randomly out of 100 visual objects in CIFAR-100 dataset. STLM01 represents the error rate on the first two visual objects, whereas STLM2 represents the error rate on the third visual object. These error rates depend on the threshold ($\theta$), and thus we measured them by varying the threshold from 0.75 to 0.9. The results with $\theta$=0.75, 0.8, 0.85 and 0.9 are shown in (A), (B), (C) and (D), respectively. The *x*-axis represents the number of examples of third-visual object stored in ML. **(E)**, The total number of examples stored in ML depending on the threshold value ($\theta$). We illustrated the mean values and standard errors calculated from 10 independent experiments, in which 3 visual objects are independently chosen; the results suggest that the variance is high. **(F)**, The error rate of STLM and LTLM. LTLM01 (STLM01) represents the error rate of LTLM (STLM) on the first two items, and LTLM2 (STLM2) represents the error rate of LTLM (STLM) on the third item. We illustrated the mean values and standard errors calculated from 10 independent experiments. All errors are estimated using the test set, and logarithmic scales are used in *y*-axis in (C) and (D).

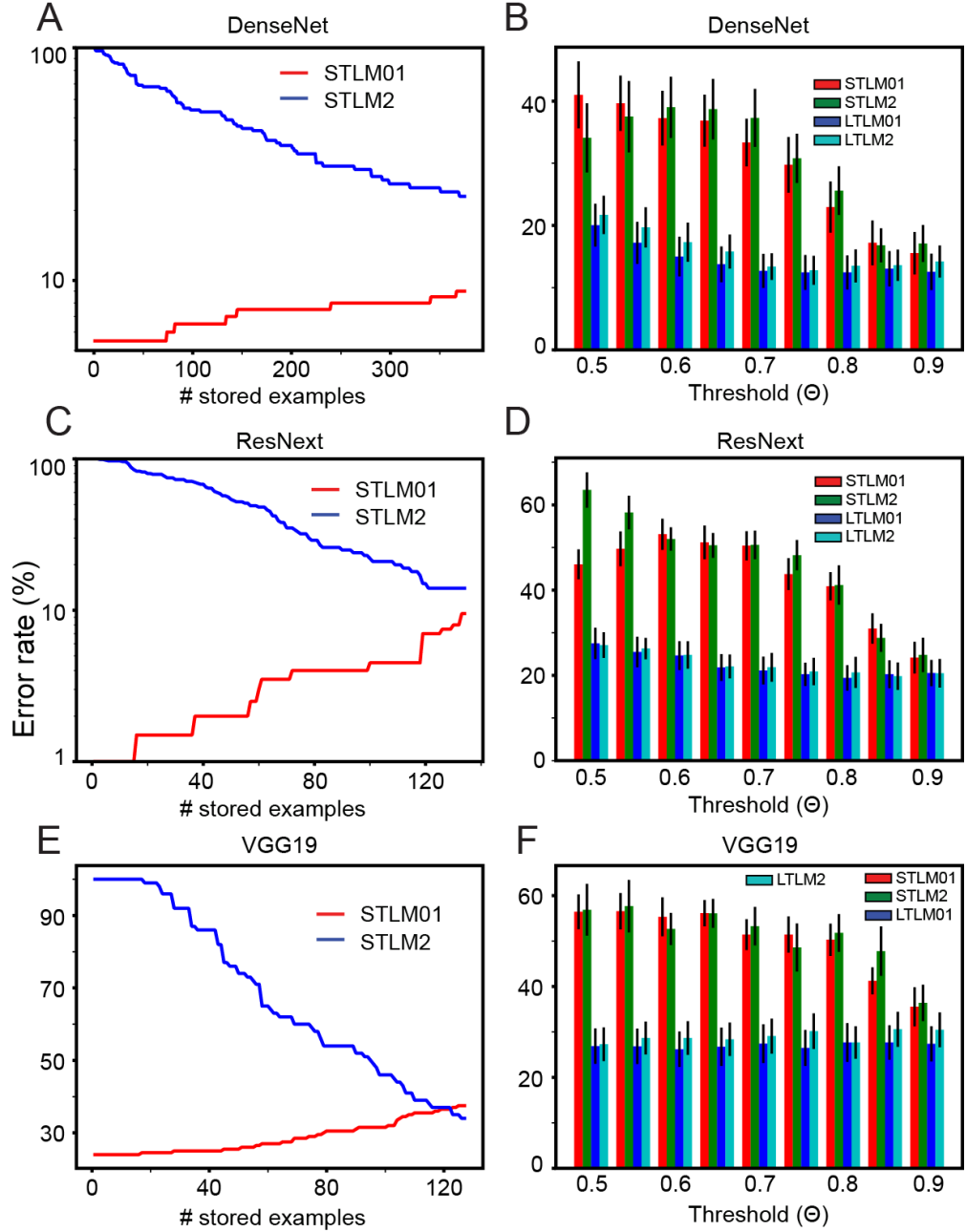

**Figure 5: DynMat's performance with alternative feature detectors**. Instead of ResNet, other convolutional networks are selected as feature detectors. **(A)**, STML's accuracy on the images of randomly chosen 3 class objects of CIFAR 100 dataset, when the feature detector is the DenseNet. Specifically, the red and blue line represent the error rate on the first two-class objects and the third-class objects during the exposure of the third class. **(B)**, The dependency of STLM's and LTLM's accuracy on the threshold θ, when the feature detector is the DenseNet. The error rate of STLM and LTLM. LTLM01 (STLM01) represents the error rate of LTLM (STLM) on the first two items, and LTLM2 (STLM2) represents the error rate of LTLM (STLM) on the third item. We illustrate the mean values and standard errors calculated from 10 independent experiments. **(C)** and **(D)**, the same as (A) and (B), but the feature detector is ResNext. **(E)** and **(F)**, the same as (A) and (B), but the feature detector is VGG19. Logarithmic scales are used in *y*-axis in (A) and (C).

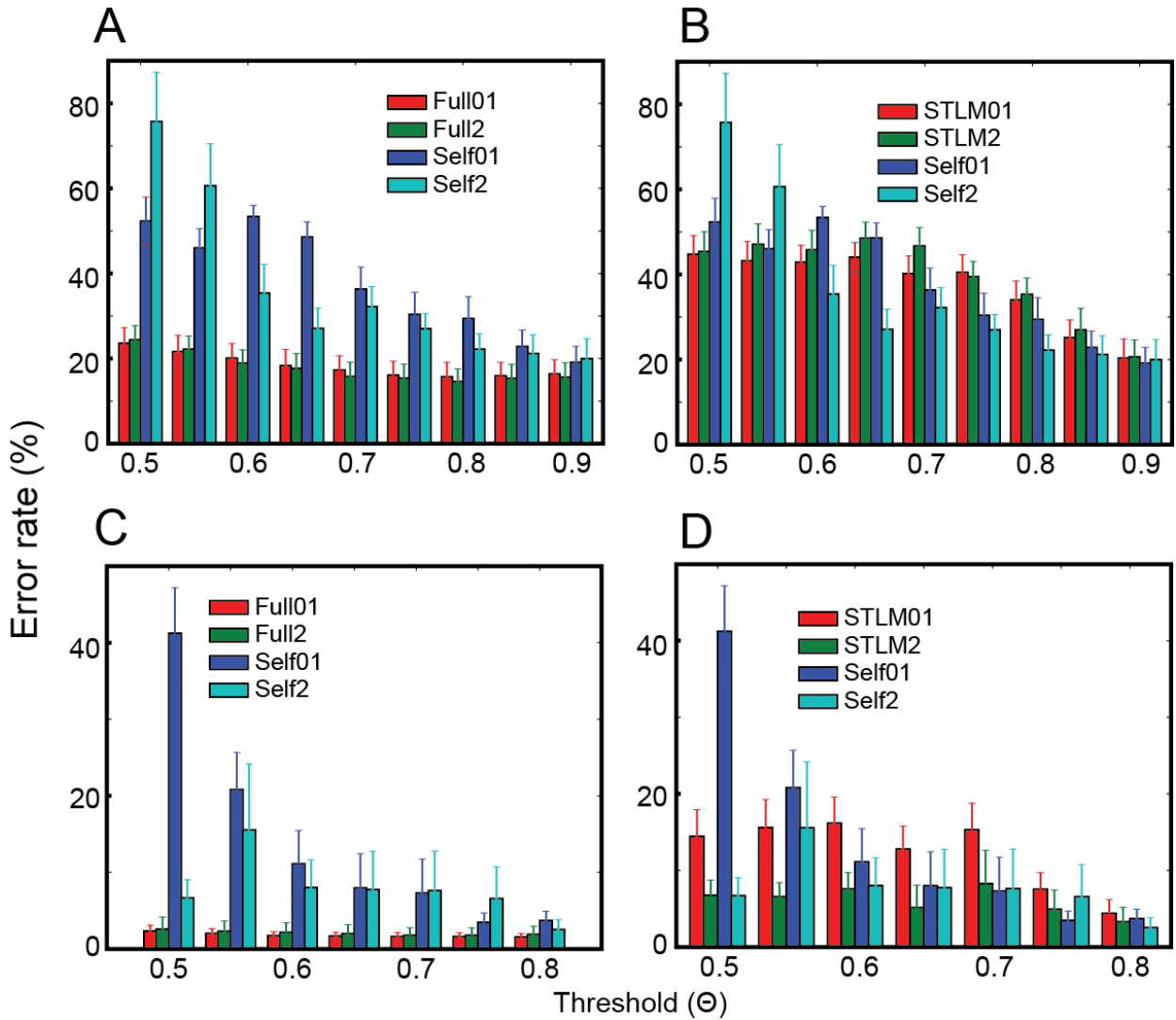

**Figure 6: The classification fidelity of LTLM. LTLM can be trained with the examples stored in ML**. We compare the classification fidelity of LTLM trained with the examples stored in ML to that of LTLM trained with the full training set, depending on the threshold ($\theta$). **(A)**, The error rates (shown in blue and cyan) of LTLM trained with examples of CIFAR-100 stored in ML and the error rates (shown in red and green) of LTLM trained with the full training set. Full01 and Full2 represent the error rate of LTLM, trained with the full training set, on the first two classes and the third class, respectively. In contrast, Self01 and Self2 represent the error rates of LTLM, trained with examples stored in ML, on the first two classes and the third class, respectively **(B)**, The comparison between error rates of LTML trained with the examples stored in ML and the error rates of STLM (depending on the threshold $\theta$). STLM01 and STLM2 represent the error rate of STLM on the first two classes and the third class, respectively. **(C)** and **(D)**, the same as (A) and (B) but DynMat is trained with fashion-MNIST, instead of CIFAR-100.

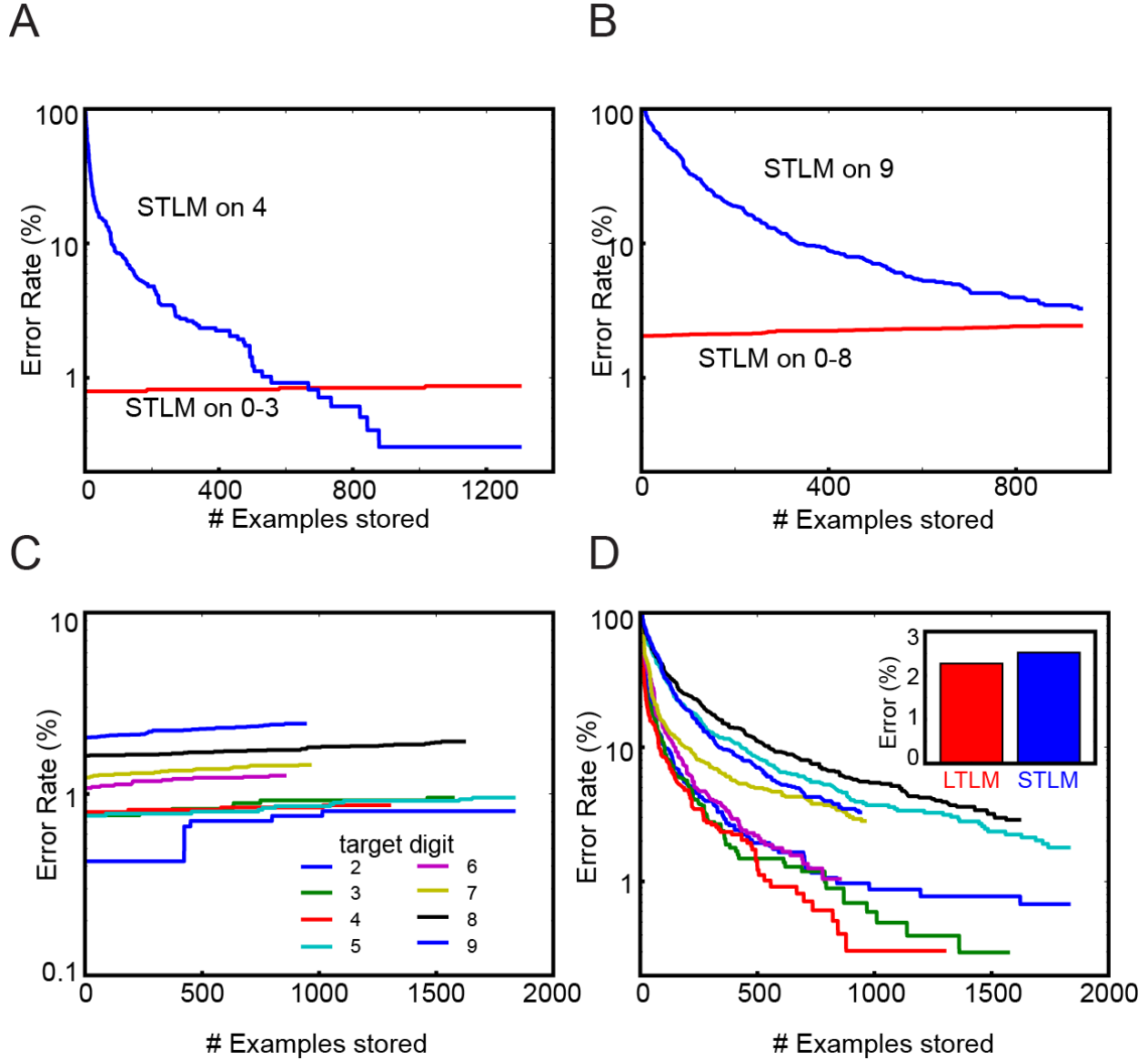

**Figure 7: STML's continuous learning of 10 digits.** 10 digits are introduced to DynMat sequentially in the ascending order. **(A)**, STLM's error rate on digit 4 in blue and four earlier digits (0-3) in red. **(B)**, STLM's error rate on digit 9 in blue and 9 earlier digits (0-9) in red. **(C)**, STML's error rates on earlier digits when new digits are introduced. The color codes represent the new digit, which is referred to as the target digit in the legend. For instance, when the target digit is 5, the cyan line represents the STML's accuracy on digits (0-4) depending on the number of examples of digit 5 stored in ML. **(D)**, STML's accuracy on the target digit. The same color codes are used as (C). That is, the cyan line represents STLM's accuracy on digit 5 depending on the number of examples of digit 5 stored in ML. The inset in (D) compares the accuracy on all 10 digits between LTLM (shown in red) and STLM (shown in blue). In all experiments, $\theta=0.8$, and 500 hidden neurons are used in LTLM. Logarithmic scales are used in *y*-axis in all panels.

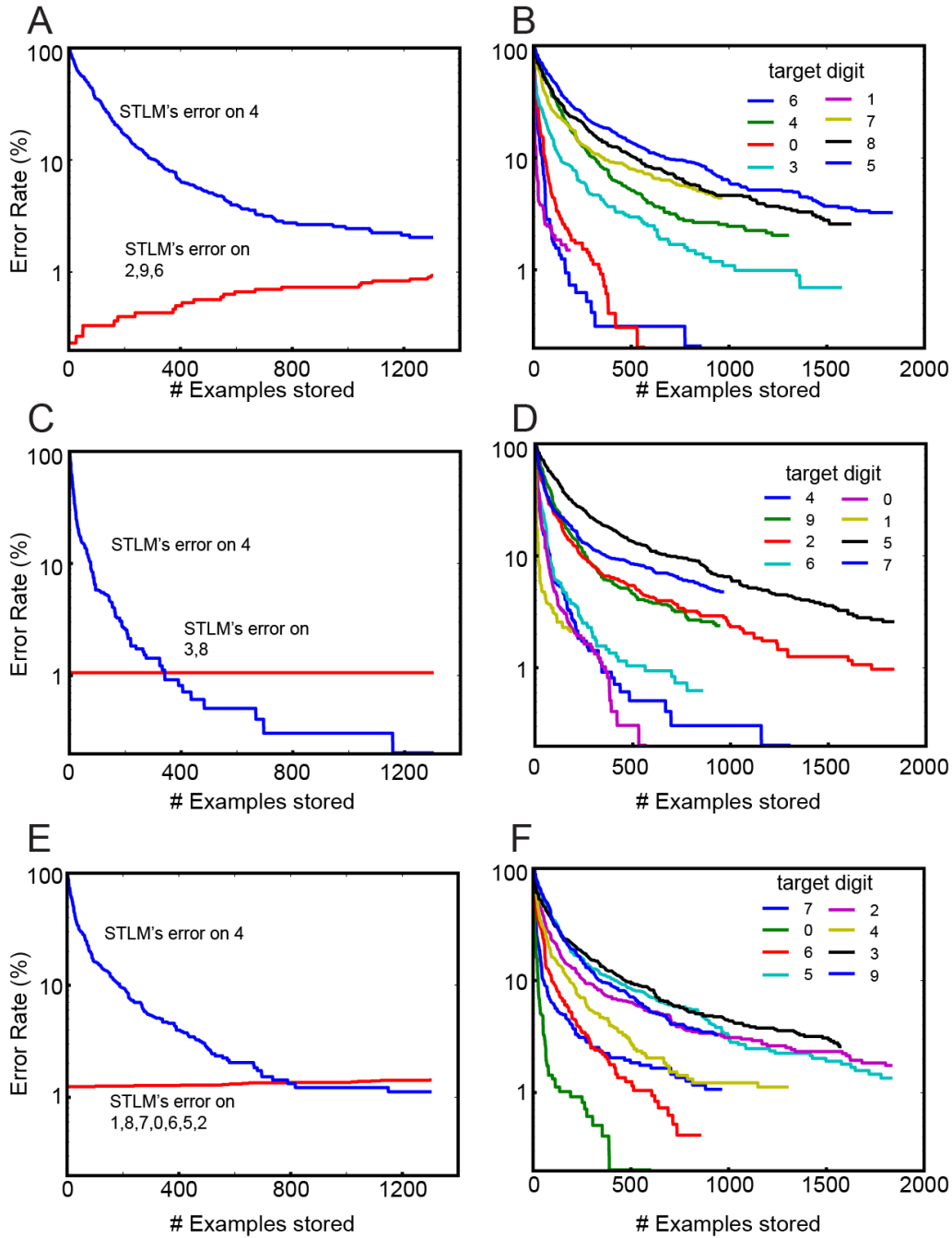

**Figure 8: STML's continuous learning of shuffled 10 digits**. We shuffle the order of 10 digits presented to DynMat. **(A)**, STLM's accuracy on the early digits (0,9 and 6) presented before digit 4. **(B)**, STLM's accuracy on new digit when it is first introduced. The order of presentation of digits are 6, 4, 0, 3, 1, 7, 8 and 5 with digits 2 and 9 initially trained. **(C)** and **(D)**, the same as (A) and (B), but the order of presentation is 4, 9, 2, 6, 0, 1, 5 and 7 with 3 and 8 initially trained. **(E)** and **(F)**, the same as (A) and (B), but the order of presentation is 7, 0, 6, 5, 2, 4, 3 and 9 with 1 and 8 initially trained. In all experiments, $\theta$=0.8. Logarithmic scales are used in $y$-axis in all panels.

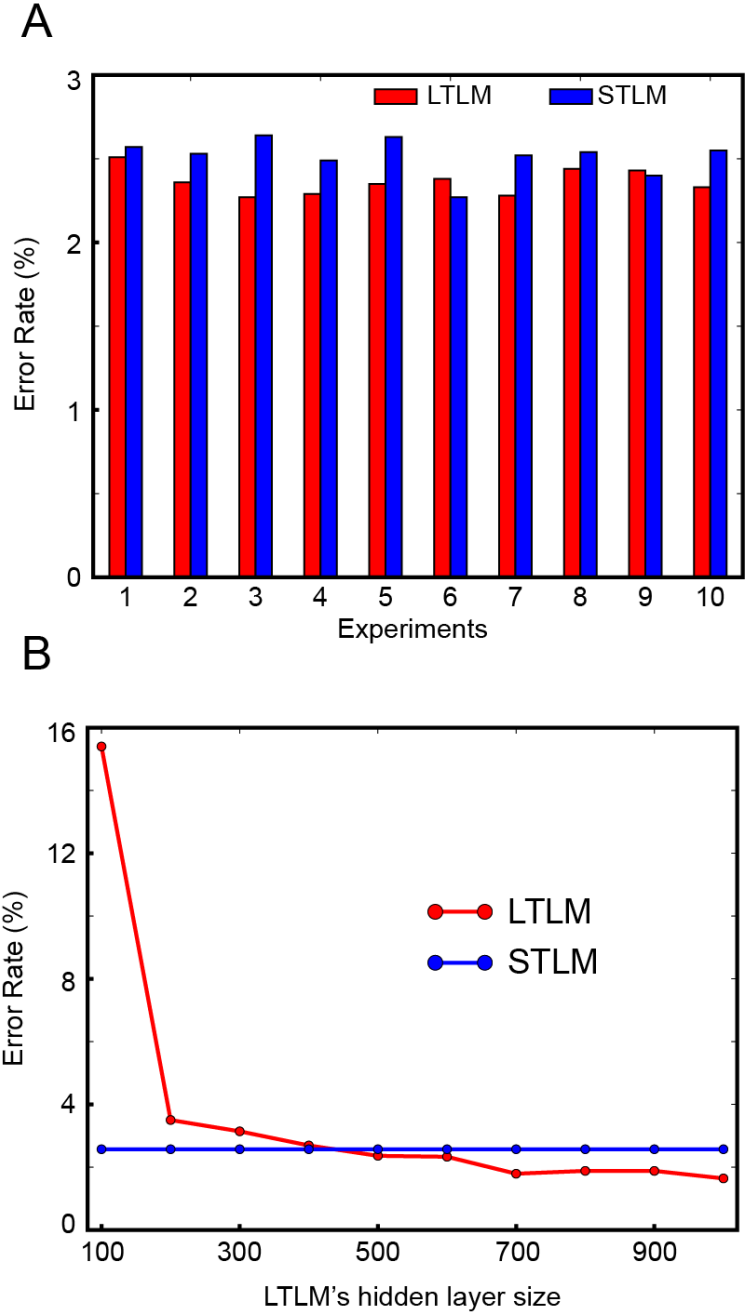

**Figure 9: LTLM's and STLM's accuracy on 10 trained digits from MNIST dataset. (A)**, The error rates in 10 independent experiments with shuffled presentations. The error rates of LTLM on 10 digits are shown in the red boxes and the error rates of STLM are shown in the blue boxes. In each experiment, the presentation orders of examples of digits are also shuffled. Due to this shuffling within the same digit, STLM stores different examples of digits across experiments, leading to small variation in STLM's accuracy. **(B)**, The comparison of LTLM's accuracy depending on hidden neurons of LTLM. Using sequences used in the experiment, we evaluate how LTLM's accuracy depends on LTLM's hidden size. Because the identical sequence is used for all experiments, STLM shows identical error rates. In all experiments, $\theta=0.8$, and 500 hidden neurons are used in LTLM.